\documentclass[conference]{IEEEtran}
\IEEEoverridecommandlockouts
\usepackage{cite}
\usepackage{amsmath,amssymb,amsfonts}
\usepackage{algorithmic}
\usepackage{graphicx}
\usepackage{textcomp}
\usepackage{diagbox}
\usepackage{bm}
\usepackage{soul}
\soulregister\ref7

\usepackage{subfigure}
\usepackage{graphics}
\usepackage{graphicx}
\usepackage{float} 
\usepackage{algorithm}
\usepackage{algorithmic}
\usepackage{amsmath}
\usepackage{xcolor}
\usepackage{multirow}
\usepackage{wrapfig}
\usepackage{setspace}
\usepackage{soul} 
\usepackage{bibspacing}

\usepackage{xcolor}
\def\BibTeX{{\rm B\kern-.05em{\sc i\kern-.025em b}\kern-.08em
    T\kern-.1667em\lower.7ex\hbox{E}\kern-.125emX}}
    
\begin{document}

\title{ Conflict-driven Structural Learning Towards Higher Coverage Rate in ATPG
}

\author{Hui-Ling Zhen$^{1}$, Naixing Wang$^{2}$, Junhua Huang$^{1}$, Xinyue Huang$^{2}$, Mingxuan Yuan$^{1}$ and Yu Huang$^{2}$\\
1. Noah's Ark Lab, Huawei; 2. Hisilicon, Huawei }




\maketitle

\maketitle

\begin{abstract}
Due to the increasing challenges posed by the relentless rise in the design complexity of integrated circuits, Boolean Satisfiability (SAT) has emerged as a robust alternative to structural APTG techniques. However, the high cost of transforming a circuit testing problem to a Conjunctive Normal Form (CNF) limits the application of SAT in industrial ATPG scenarios, resulting in a loss of test coverage.
In Order to address this problem, this paper proposes a conflict-driven structural learning (CDSL) ATPG algorithm firstly, in which the conflict-driven heuristic methods in modern SAT solver are implemented on the logic cone of fault propagation and activation directly. The proposed CDSL algorithm is composed of three parts: (1) According to the implication graph, various conflict constraints have been learned to prune search space. (2) Conflict-driven implication and justification have been applied to increase decision accuracy and solving efficiency. (3) A conflict-based diagnosis method is further proposed in the case of low coverage debug, leading to making the aborted faults testable by relaxing or modifying some constraints on primary inputs. Extensive experimental results on industrial circuits demonstrate the effectiveness and efficiency of the proposed CDSL algorithm. It is shown that compared with the SAT-based ATPG, the proposed CDSL can on average decrease $25.6\%$ aborted faults with $94.51\%$ less run time. With a two-stage computational flow, it has shown that the proposed CDSL can lead to $46.37\%$ less aborted faults than a one-stage structural algorithm, further with the $3.19\%$ improvement on fault coverage. In addition, the conflict diagnosis can lead to $8.89\%$ less aborted faults on average, and $0.271\%$ improvement in fault coverage rate.

\end{abstract}


\section{Introduction} \label{sec:intro}

Continuous progress in decreasing device sizes and increasing design complexity has brought increasing demand for high product quality and low defective parts-per-million (DPPM) goals. Thus, scan-based structural testing has become even more important than ever, and Automatic Test Pattern Generation (ATPG) has served as an essential procedure for generating appropriate test patterns for testing logical faults that model physical defects.

Given a targeted fault of the circuit-under-test, the goal of ATPG is to either generate a test pattern for the targeted fault (i.e., finding the test vector that can differentiate the good and faulty machines and that such fault is detectable) or prove that it is undetectable (i.e. there is no test vector that can differentiate the good and faulty machines). There have been several structural algorithms for ATPG, such as D-algorithm~\cite{dalg} and its advanced variants~\cite{tea,socrates}.

There are two core problems in ATPG. One is how to improve decision efficiency under a given backtrack limit, especially considering a large number of hard-to-detect faults in today's complex designs. There mainly exist two methods to solve this problem. One is to utilize Boolean Satisfiability (SAT) solver directly~\cite{incresat, recentadvances}. Unlike structural ATPG working on a circuit network, SAT-based ATPG makes use of symbolic calculation techniques to implement efficient conflict-driven search on the Conjunctive Normal Form (CNF). Many SAT-based ATPG algorithms have been proposed, such as TG-Pro~\cite{tg-pro}, TIGUAN~\cite{tiguan}, and PASSAT~\cite{passat}. Similar SAT-based techniques have been applied, so as to insert test points for low-capture-power testing while maintaining the same fault coverage~\cite{formal-sat}.

\begin{wrapfigure}{l}{0.25\textwidth}
\centering
\begin{minipage}{8cm}
\includegraphics[width = 0.56\textwidth] {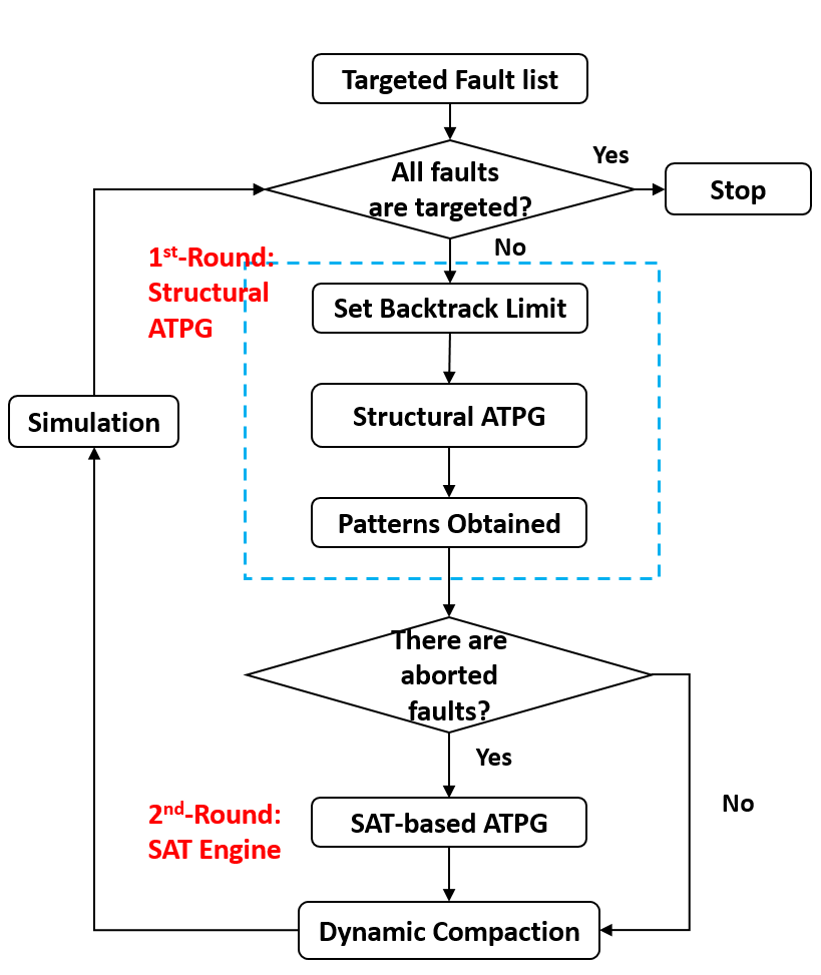}
\end{minipage} 
\caption{\small A hybrid computational flow in ATPG, which begins at the structural ATPG and ends with the SAT.  }
\label{fig:current-flow}
\end{wrapfigure} 
A hybrid computational flow composed of structural ATPG and SAT-based algorithms has been proposed, as shown in Figure~\ref{fig:current-flow}~\cite{incresat}. Here, the structural ATPG algorithm is adopted firstly under a given backtrack limit and it targets relatively easy-to-detect faults, which can be detected via a test pattern or proved to be undetectable. Then SAT targets the hard-to-detect faults which are aborted by the structural ATPG. Unlike structural ATPG, which is performed directly on the circuit, SAT-based algorithms rely on the CNF transformed from the logic cone of fault propagation and activation. This transformation is an extra step in SAT-based algorithms.

\begin{figure}[htbp]
\centering
\begin{minipage}{8cm}
\centering
\includegraphics[width = 0.8\textwidth] {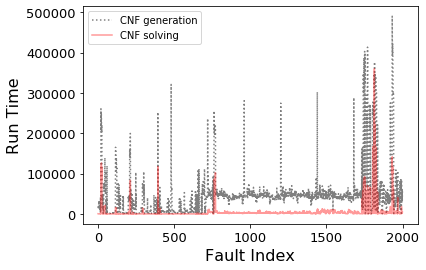}
\end{minipage} 
\caption{\small Comparison between the CNF generation time vs. solving time (in CPU microseconds). The horizontal axis is the fault index, while the vertical axis is the respective runtime.  }
\vspace{-0.3cm}
\label{fig:compare_sol_gene}
\end{figure} 
Take one circuit $Stuck~4$ as an example (with additional details provided in Section~\ref{sec:experiments}), we demonstrate a significant challenge for SAT in Figure~\ref{fig:compare_sol_gene}. The figure examines the time required for the transformation from the circuit to CNF in comparison to the related solving time. All targeted faults are stuck-at, and the SAT-based framework follows TG-Pro~\cite{tg-pro}. The chosen SAT Solver is Kissat~\cite{kissat}, a reference SAT solver in SAT competition 2022. It is revealed that the transformation process requires more runtime than solving itself. This indicates that despite the significant advancements made in SAT solver, which have displayed considerable potential in solving ATPG problems~\cite{recentadvances}, the additional overhead required for CNF transformation limits SAT's applications in industrial ATPG. 
Several works have been done to alleviate this problem. An incremental SAT-based framework has been proposed which aims to generate CNF incrementally and find the final solutions, or prove UNSAT, via partial CNF, hence decreasing the transformation time as well as solving time on average~\cite{incresat}. Preprocessing methods have been utilized to simplify the logic cone of fault propagation and activation, leading to a decrease in the generation and solving time by solving an equivalent substitute~\cite{incresat2}.

Nevertheless, the CNF transformation on large-scale circuits remains a big bottleneck, resulting in utilizing SAT solver being directly limited. Thus, the second method is to attempt to utilize SAT's heuristics on the circuit. A conflict-driven recursive learning which combines with a dynamic decision ordering technique has been proposed to resolve hard-to-resolve faults~\cite{conflict3}. A conflict-driven implication method has been proposed to improve the justification efficiency and avoid the over-specifications of test vectors~\cite{conflict1}. An untestable debug algorithm has also been utilized for low test coverage analysis~\cite{conflict2}. However, the method of constructing learning conflicts in modern SAT solvers, like the unique implication point (UIP), has not been considered.

The other problem is that the ATPG constraints are usually conservative during the early stage of the design~\cite{conflict2}. The conservatism often results in the implementation not being sufficiently mature in practice. Therefore, in the early stages, the DFT engineers have some degree of freedom to relax or modify certain constraints, making that some of the aborted faults as well as untestable faults which are not led by the circuit structure can be potentially resolved.
To address this issue, we employ a conflict diagnosis approach after running ATPG engine to resolve low test coverage. Take one aborted fault as an example. We consider that the reason for abortion is due to the encountered conflicts exceeding the backtrack limit. Finally, the statistical analysis for the learnt conflicts would provide meaningful suggestions to the DFT engineer, leading to a decrease in the number of aborted or untestable faults and improving the coverage rate.

Motivated by the aforementioned discussions, this paper proposes a conflict-driven structural learning (CDSL) ATPG algorithm, which aims to utilize the advantages brought by the structural ATPG and SAT-based algorithms. To summarize, our contributions include: 

(i) \textbf{We first build learnt conflict-based constraints directly on circuit,} aiming to prune the searching space by using the optimization process data. According to the implication graph which is directly related to the decision-making process, we construct two kinds of conflict constraints, i.e., decision-variable-based constraint and UIP-based constraint, leading to avoiding meaningless searching in subsequent iterations. 

(ii) \textbf{We adopt the conflict-driven decision rules to improve the decision accuracy.} After accumulating the learnt conflicts, we construct new implications and justification approaches based on those conflicts. Extensive experiments demonstrate the effectiveness of conflict constraints on implication efficiency with fewer backtracks and aborted faults. 

(iii) \textbf{We further construct the conflict diagnosis according to the learnt conflicts in the case of low coverage debug.} In this method, we utilize the learnt conflicts to analyze the reason from PIs' constraints and relax or modify certain of them, aiming at further improving the test coverage rate.  

The remainder of this paper is organized as follows. After some preliminaries in Section~\ref{sec:formulation}, Section~\ref{sec:algorithm} presents our new SAT-based ATPG approach. Experimental results are demonstrated in Section~\ref{sec:experiments}, in which we show the effectiveness of the proposed framework both on solution quality and runtime. Finally, we conclude this work in Section~\ref{sec:conclusion}.

\section{Preliminaries}\label{sec:formulation}

\subsection{Conflict-Driven-Clause-Learning (CDCL) in SAT}

SAT-based ATPG makes use of efficient SAT solvers to solve APTG problems. It begins with building a CNF format SAT model which represents the fault condition and propagation between the PIs and the POs. In general, a CNF formula $\phi$ consists of a conjunction of clauses $\omega$, each of which denotes a disjunction of literals. A literal is either a variable $x_i$ or its complement. Each variable can be assigned a logic value, either $0$ or $1$. Any general Boolean problems can be represented as a CNF formula model. A SAT solver either finds an assignment such that $\phi$ is satisfied, or proves that no such assignment exists, i.e., UNSAT. A key heuristics in modern SAT solver is Conflict-Driven-Clause-Learning (CDCL) algorithm~\cite{recentadvances}. In general, CDCL is a Branch-and-Bound (BB) search framework, in which each step a literal and a propositional value (either 0 or 1) are selected for branching purposes. A key characteristic of CDCL is to learn new clauses from conflicts during backtrack searches. 


\subsection{Structural ATPG Algorithm}

Different from SAT-based algorithms, the structural ATPG algorithm is performed on the circuit directly. Until now, several kinds of algorithms like D-algorithm, PODEM, and FAN have been proposed. In practice, D-algorithm tries to propagate the stuck-at-fault value denoted by $D$ (for Stuck-at-0) or $\overline{D}$ (for Stuck-at-1) to a primary output (PO)~\cite{dalg}. The conventional D-algorithm generates a decision structure to evaluate the value of every node in the circuit to obtain the test vectors. PODEM and FAN are the advanced variants by limiting the searching space and accelerating backtracing, while PODEM limits the searching space only to Primary Inputs (PIs)~\cite{podem}, and FAN limits the searching space to headlines~\cite{fan}. 


\subsection{Comparison between Structural ATPG and CDCL}

\begin{wrapfigure}{l}{0.24\textwidth}
\centering
\begin{minipage}{8cm}
\includegraphics[width = 0.52\textwidth] {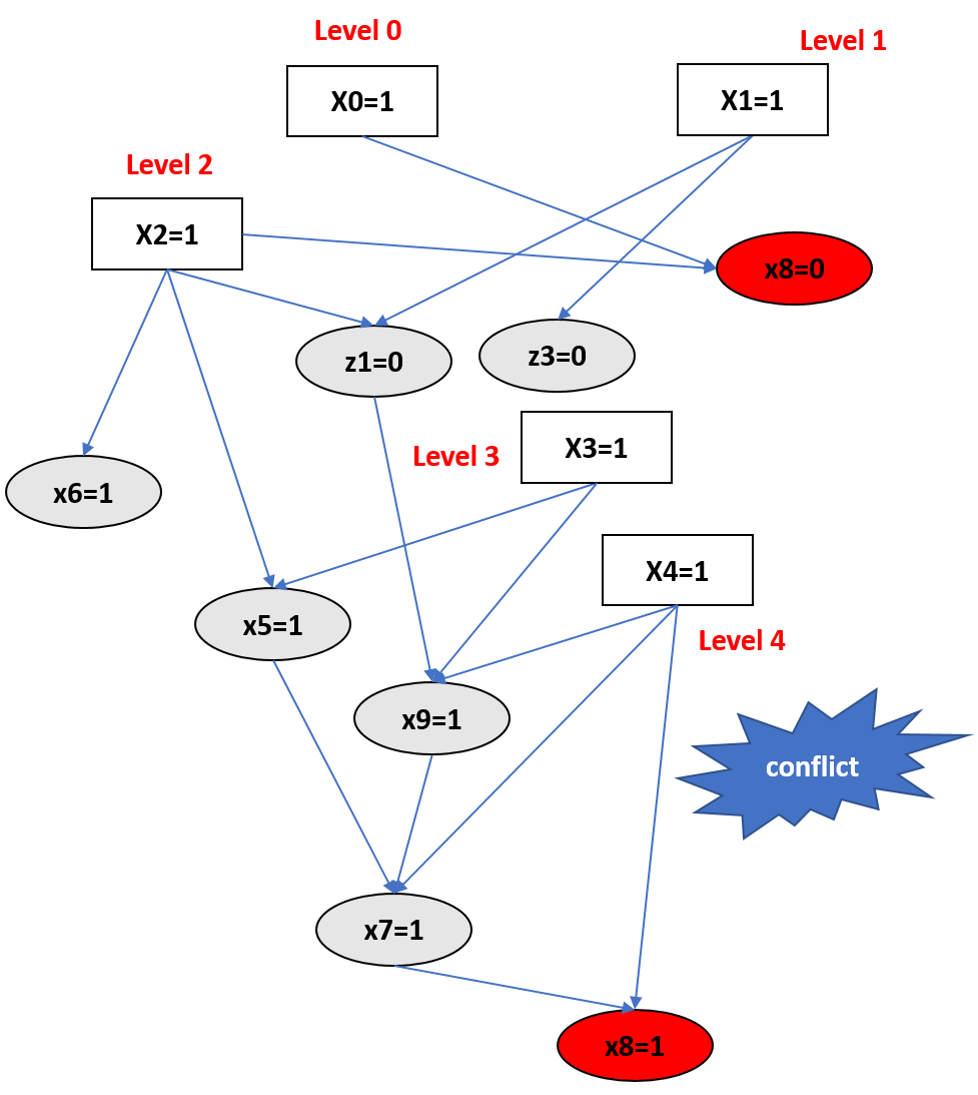}
\end{minipage} 
\caption{\small An example of a decision-making process. All decision variables are in square boxes, and implications in are in oval boxes. The related decision level is also labeled.   }
\vspace{-0.3cm}
\label{fig:example-dec}
\end{wrapfigure}

There exists a certain difference between CDCL and structural ATPG algorithm. The first difference must root in the branching rules. The structural ATPG algorithm is requirement-driven~\cite{dalg}, which denotes that the decision order accords with the fault propagation and circuit structural characteristics. Unlike this, the initial decision order in CDCL accords to the input literal order which is random, and this order is modified based on the literal's frequency in learnt conflict constraints after some backtracks. The second difference roots the backtrack rules after conflict occurs. We take an example to discuss other differences, as shown in Figure~\ref{fig:example-dec}. All the decision variables ($x0$, $x2$, $x3$, and $x4$) are in square boxes, while all the implicated variables are in oval boxes. Each decision variable is assigned with a decision level according to the decision order. The direction of the arrow is consistent with the direction of the implication. 

Figure~\ref{fig:example-dec} shows that, after the fourth decision variable, a conflict occurs (i.e., $x8$ cannot be $0$ and $1$ at the same time). In the structural ATPG algorithm, the decision pointer will backtrack to the last decision variable (i.e., $x3$), but without analysis of the reason for the occurrence of conflicts. In the given conflict-driven methods~\cite{conflict1,conflict2,conflict3}, there will be added one learnt conflict constraint $x4 \neq 1$, which limits the following implications under new searching rules. 
Apparently, a better searching strategy must combine both advantages of structural ATPG and CDCL, i.e., the branching rules follow the structural ATPG algorithm which aims to decrease the cost of wrong decisions, while once conflict occurs, the reasons for conflict should be considered like CDCL to avoid same wrong searching path.


\section{Proposed CDSL Algorithm} \label{sec:algorithm} 

\begin{figure}[htbp]
    \centering
    \begin{minipage}{9cm}
    \includegraphics[width = 0.97\textwidth] {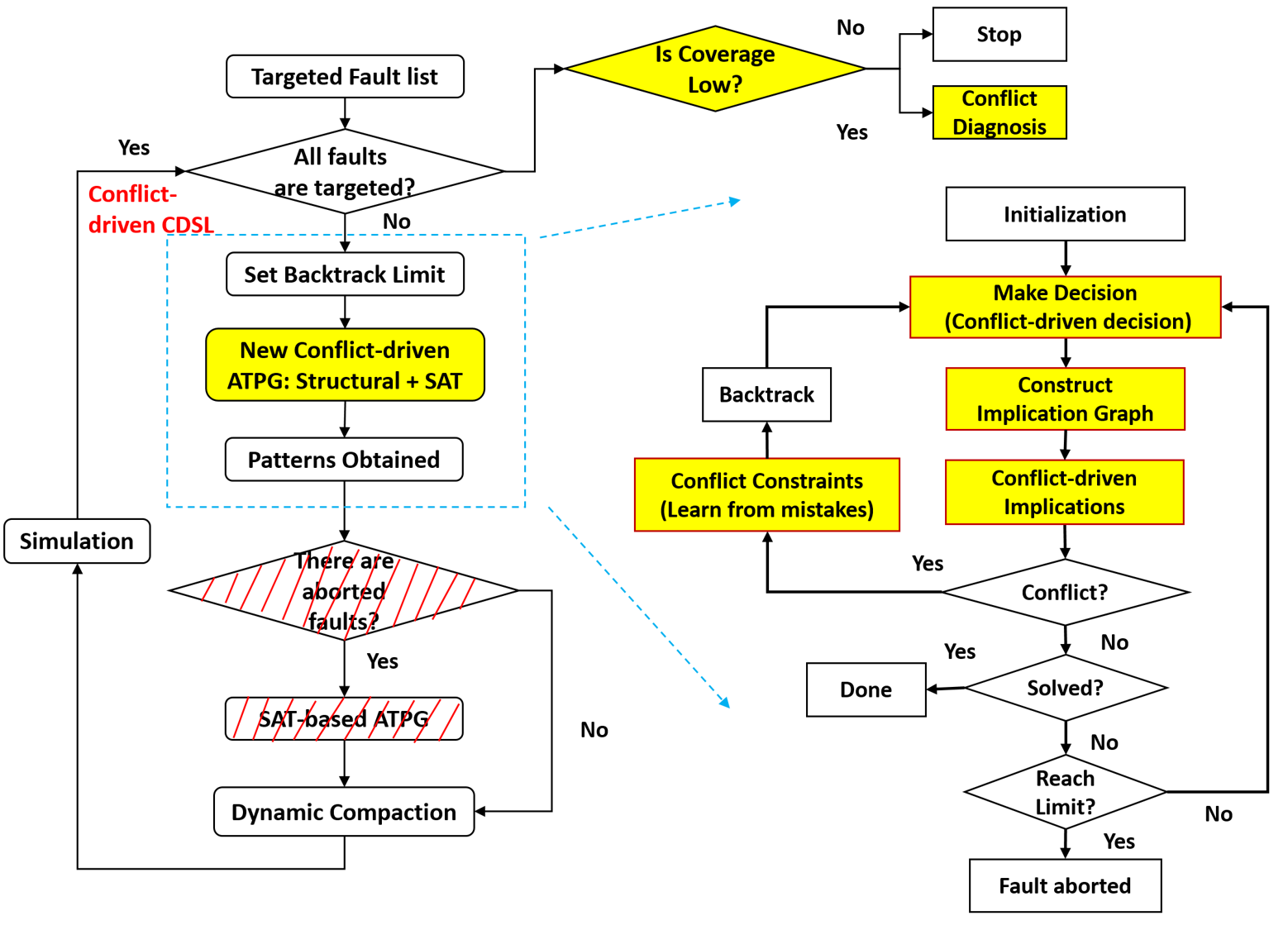} 
    \end{minipage}
    \caption{\small New proposed CDSL algorithm. Different from the conventional structural ATPG algorithm, we incorporate SAT's heuristics such as learnt conflict constraints, conflict-driven implication, and conflict-driven branch/decision, aiming to prune the searching space based on data from the optimization process and find solutions or prove UNSAT, with fewer backtracks. After the new ATPG computation, we propose to add the conflict diagnosis in case of low coverage.  }
    \label{fig:new_flow}
\end{figure}

Considering the above, we propose a conflict-driven structural learning (CDSL) ATPG algorithm which combines two methods, as shown in Figure~\ref{fig:new_flow}. 
Compared with the conventional structural ATPG and SAT-based ATPG algorithms, the CDSL algorithm has two advantages: (1) It accumulates conflict constraints after backtracks, with the aim of avoiding the same wrong decisions and finding solutions with fewer backtracks. (2) It employs conflict-driven implications to prune the searching space and conflict-driven branching rules, with a score heuristics, to improve decision accuracy.

Given a fault site, we first trace the circuit to get the logic cone related to fault propagation and activation. The decision rules begin at the fault site and follow the conventional structural ATPG algorithm until one conflict occurs. In the process, all structural ATPG algorithms like D-algorithm, PODEM, and FAN can be used. 

\subsection{Implication Graph}

Firstly, we construct an \textit{implication graph} according to the decision-making process: 

(1) We construct a directed acyclic graph in which each vertex represents a variable's assignment, and each incident edge to a vertex represents the reason leading to that assignment. If one implication is inferred via other implications, there also exists an edge among different implications. Thus, decision variables have no incident edges in contrast to implied variables that have assignments forced during propagation.  

(2) Each decision variable is assigned a decision level according to the related decision-making order, while its related implications have the same decision level. 

Note that each variable in CDSL's implication graph denotes a logic gate. Once a conflict occurs, the proposed CDSL algorithm would trace the implication graph to find all the historical assignments which result in the conflict and construct learnt conflict constraint.

\subsection{Learnt Conflict Constraints} 

Take Figure~\ref{fig:example-dec} as an example, in which a conflict occurs through $x_8$, we construct two kinds of learnt conflict constraints in the proposed CDSL algorithm.

\textbf{(1) Decision Variable-based Conflict.} The basic principle is that the current conflict, at least, is caused by all historical decision variables. As shown in Figure~\ref{fig:example-dec}, before the conflict occurs, there are four decision variables, i.e., $x_0=1$, $x_1=1$ $x_2=1$, $x_3=1$ and $x_4=1$, thereby we can add a learnt conflict constraint as $x_0 + x_1 + x_2 + x_3 + x_4$ that is constructed via the decision variables. It denotes that in the following decision-making process, even though the four variables can serve as decision variables, they cannot repeat the same assignments, in other words, when it is found that three of these variables repeat the historical assignments, the fourth variable must take the opposite assignment.

\textbf{(2) Unique Implication Point (UIP)-based Conflict.} A UIP is a special node that any node at the current decision level such that any path from the decision variable to the conflict node must pass through it~\cite{uip_sat}. As shown in Figure~\ref{fig:example-dec}, the conflict occurs in node $x_8$ whose decision level is $4$. The inference of UIP-based learnt conflict constraints can be given as follows: 

(i) We first find the direct reason for the conflict node. Figure~\ref{fig:example-dec} exhibits that one $x_8$'s direct reasons are $x_4$ and $x_7$, and the other $x_8$'s direct reason is $x_0$ and $x_2$. Hereby, both $x_0$, $x_2$, and $x_4$ are decision variables and their decision level is $0$, $2$, and $4$, respectively. $x_7$ is implications from $x_4$ , $x_5$, and $x_9$. Thus, the direct learnt conflict constraint can be given as $x_0 + x_2 + x_4 + x_7$. 

(ii) Check the decision level, and we should decide whether some of the reason nodes are replaced by the corresponding parents. The evaluation rule is that in the final learnt conflict constraint, there exists only one variable whose decision level is the same as the conflict node, and this variable is UIP. 

(ii-a) Consider $x_0 + x_2 + x_4 + x_7$, since both $x_7$, $x_9$, and $x_4$ are in decision level $4$ and $x_4$ is a decision variable, we utilize $x_7$'s parent nodes (i.e., $x_4$, $x_5$ and $x_9$) to replace it. After deduplication, the learnt conflict constraint is updated as $x_0 + x_2 + x_4 + x_5 + x_9$, in which the decision levels of $x_5$ and $x_9$ are $3$ and $4$, respectively. 

(ii-b) Since $x_9$ and $x_4$ are in the same decision level, we utilize $x_9$'s parents (i.e., $z_1$, $x_3$ and $x_4$) to replace it, and then the learnt conflict is updated as $x_0 + x_2 + x_4 + x_5 + z_1 + x_3$.

Finally, we can obtain the UIP-based learnt conflict constraint as $x_0 + x_2 + x_4 + x_5 + z_1 + x_3$. Considering that the only variable whose decision level is the same as the conflict node is $x_4$, thus, $x_4$ serves as the UIP node. Note that we only show the learnt relationship among different variables, not including the logic values. After accumulating different learnt conflict constraints, the proposed CDSL algorithm will utilize those in the following three aspects:

\subsection{\textbf{Conflict-driven Implications}}

All learnt conflict constraints are applied for the implication phase, aiming to avoid repeating the wrong searching paths. Take the UIP-based learnt conflict constraint $x_0 + x_2 + x_4 + x_5 + z_1 + x_3$ of Figure~\ref{fig:example-dec} as an example, if we find that five of the related variables (i.e., $x_0$, $x_2$, $x_4$, $x_5$ and $z_1$) have the same assignments with historical ones, the sixth must be assigned as the opposite value. To avoid the extra computational overhead when too many learnt conflict constraints are accumulated, we also add a forgotten rule in the implication phase: if one learnt conflict constraint is not utilized in recent $N$ loops, this constraint is considered to be no longer relevant and it would be deleted in the following loops. Hereby, $N$ is a hyperparameter. 

\subsection{\textbf{Conflict-driven Branch Heuristics}}

The learnt conflict constraints can also be applied through Variable State Independent Decaying Sum (VSIDS) heuristic, aiming to improve the decision accuracy in the following decision phase. There are three steps in the VSIDS strategy: 

a) We start by assigning each variable a floating point score. When a conflict occurs, the activity of some variables is increased by $1$. In general, the initial score is set to $0$. 

b) After each conflict, the variable activity is decayed periodically, aiming to trade off the historical decisions and following ones. Such decay factor is set $[0,1]$. 

c) To balance VSIDS and structural strategies, we would check each variable's score during branching. The variable with the highest score is selected under a given probability. 

Further, different from the structural ATPG algorithm which requires backtracking to the last decision variable, we adopt a \textbf{ non-chronological backtrack} rule in the proposed CDSL algorithm. This rule accords with the UIP-based conflict constraint, and the backtrack point is the variable that is with the largest decision level except for the UIP node. Take Figure~\ref{fig:example-dec} as an example, the scores of $x_0$, $x_5$, $x_3$ and $x_4$ are higher than others' after both decision-variable-based and UIP-based conflict constraints are accumulated, and once one conflict occurs, the backtrack point is chosen as $x_3$. 

\subsection{Conflict Diagnosis for Low Coverage Debug}\label{subsec:diagnosis}

Except for the implications and branching, we also explore adopting the conflict diagnosis to beat the low test coverage in the initial phase of design: 

(i) Compute each logic gate's score according to the frequency in the learnt conflict constraints.

(ii) Choose the top-k gates according to the score's rank. Then trace the circuit to find the related external constraints. Usually, those constraints are put on either primary inputs or the fan-in gates of decision level $0$. 

In conflict diagnosis, we choose to relax or modify the identified external ATPG constraints, which would provide an opportunity to make the aborted or untestable fault testable. 


\section{Experimental Results}\label{sec:experiments}

\subsection{Experiments Setup}

In this section, we aim to evaluate the proposed CDSL algorithm from the following three aspects:  

\begin{itemize}
    \item[\textbf{RQ1}:] Can it have a performance advantage over the traditional SAT-based algorithms? 
    
    \item[\textbf{RQ2}:] Can it be beneficial for improving test coverage compared to the structural algorithm? 
    
    \item[\textbf{RQ3}:] Can the conflict diagnosis be exploited to debug the aborted or untestable faults? 
\end{itemize} 
In the following, the CDSL framework is implemented on the structural D-algorithm. and its performance is evaluated from two perspectives, one is the number of aborted faults (unobserved faults, abbreviated as UO) under the set aborted limit, the other one is fault coverage rate, i.e., $\textit{Fault Coverage} =\frac{N_{Testable}}{N_{Total}}$, where $N_{Total}$ and $N_{Testable}$ are the number of total faults and testable faults, respectively. All experiments are carried out for industrial circuits, and their designs are shown in Table~\ref{table:design_chrac}.

\begin{center}
\begin{table}[htbp]
\centering
\vspace{-0.6cm}
\begin{minipage}{7.1cm}
\centering
\caption{Design Characteristics }
\label{table:design_chrac}
\def\arraystretch{1.5}\tabcolsep 2pt
\resizebox{\textwidth}{22mm}{
\begin{tabular}{c|c|c|c|c|c|c|c}
\hline 
Circuit & Fault Type & $\#$gates & $\#$State & Circuit & Fault Type &  $\#$gates & $\#$State   \\ 
\hline
Stuck~1 & Stuck-at & 246078 &  14979 & Tran~1 & Transition & 139871 & 9644    \\
\hline 
Stuck~2 & Stuck-at & 246078 & 14979 & Tran~2 & Transition & 785559	& 26288   \\
\hline
Stuck~3 & Stuck-at & 221004 & 18190 & Tran~3 & Transition &  785559 & 383963 \\
\hline
Stuck~4 & Stuck-at & 78600 & 12047 & Tran~4 & Transition &  785559 & 357483 \\
\hline
Stuck~5 & Stuck-at & 221004 & 18190 & Tran~5 & Transition & 221004 & 357483 \\
\hline
Stuck~6 & Stuck-at & 206221 & 15772 & Tran~6 & Transition & 221004 & 331291 \\
\hline
Stuck~7 & Stuck-at & 56586 & 8194 & Tran~7 & Transition & 221004 & 374009  \\
\hline
Stuck~8 & Stuck-at & 221004 & 357483 & Tran~8 & Transition & 206221 & 331291 \\
\hline
Stuck~9 & Stuck-at & 246078 & 331291 & Tran~9 & Transition &  206221 & 331291 \\
\hline
Stuck~10 & Stuck-at & 785559 & 26288 & Tran~10 & Transition & 221004 & 331291 \\
\hline 
\end{tabular}}
\end{minipage}
\end{table}
\vspace{-0.3cm}
\end{center}


\subsection{Evaluation on Run Time}

To answer \textbf{RQ1}, we choose stuck-at faults to compare the proposed CDSL with SAT-based methods, as shown in Table~\ref{table:com_uo_sat}. The first column is the circuit name. The second and third columns show the number of aborted faults led by the proposed CDSL algorithm and related run time (in CPU seconds), respectively. Hereby, the aborted limit is set as $100$. Then from left to right, there are four different baselines to evaluate the CDSL algorithm: 

i) A basic SAT-based framework, TG-Pro~\cite{tg-pro}. It is also the latest open-source framework. The SAT solver is chosen as Kissat2022~\cite{kissat}. 

ii) The basic D-algorithm. It is also a module of the proposed CDSL algorithm. 

iii) An incremental SAT-based ATPG method with preprocessing procedure~\cite{incresat}.

iv) A SAT-based ATPG method with a fault analysis module~\cite{genesat}, which is a trained neural network and predicts the fault classification for appropriate algorithm selection. 

It is shown that compared with the conventional SAT-based ATPG and structural D-algorithm, the proposed CDSL algorithm can decrease the aborted faults by $25.6\%$ and $49.88\%$ on average, while the run time is decreased by $94.51\%$ and $25.88\%$, respectively. Although the two new variants, i.e., the SAT-based ATPG with preprocessing or with the learnt network-based fault analysis can lead to fewer aborted faults and better run time, the proposed CDSL can also decrease the UO by $45.23\%$ and $12.35\%$, respectively, and the related run time can be decreased $58.79\%$ and $93.09\%$.


It is worth mentioning that when the backtrack limit is the same, both the conventional structural ATPG and the proposed CDSL algorithm can lead to fewer aborted faults than SAT-based methods. It is because the SAT's heuristics, such as branching, restart, and local search, totally rely on the score based on accumulated conflicts. It denotes that the limited conflict constraints may affect the performance of heuristics.

\begin{center}
\begin{table}[htbp]
\centering
\vspace{-0.2cm}
\begin{minipage}{8.9cm}
\centering
\caption{Performance of CDSL on UO and Run Time }
\label{table:com_uo_sat}
\def\arraystretch{1.5}\tabcolsep 2pt
\resizebox{\textwidth}{32mm}{
    \begin{tabular}{|l|l|l|l|l|l|l|l|l|l|l|}
    \hline
\multirow{2}{*}{Circuit}& \multicolumn{2}{|c|}{ CDSL } &    \multicolumn{2}{|c|}{ TG-Pro } & \multicolumn{2}{|c|}{ Structural } & \multicolumn{2}{|c|}{ Incre } & \multicolumn{2}{|c|}{Neural } \\
\cline{2-11} & UO & time   & UO & time &  UO & time  & UO & time & UO & time  \\ \hline
        Stuck~1 & 147 & 229 & 174 & 10952 & 226 & 814 & 162 & 1528 & 162 & 9125  \\ \hline
        Stuck~2 & 352 & 167 & 559 & 1722 & 793 & 128 & 638 & 218 & 475 & 1522  \\ \hline
        Stuck~3 & 253 & 33 & 195 & 780 & 271 & 58 & 139 & 678 & 175 & 672 \\ \hline
        Stuck~4 & 1 & 53 & 7 & 1103 & 8 & 101 & 12 & 206 & 7 & 856  \\ \hline
        Stuck~5 & 144 & 18 & 119 & 393 & 158 & 36 & 105 & 79 & 110 & 326  \\ \hline
        Stuck~6 & 1343 & 365 & 1318 & 5165 & 1949 & 1307 & 2125 & 806 & 986 & 4238  \\ \hline
        Stuck~7 & 236 & 97 & 485 & 1389 & 453 & 92 & 383 & 234 & 429 & 1109  \\ \hline
        Stuck~8 & 601 & 550 & 518 & 10543 & 664 & 498 & 836 & 631 & 492 & 7692  \\ \hline
        Stuck~9 & 514 & 75 & 987 & 977 & 1303 & 812 & 1189 & 235 & 836 & 901 \\ \hline
        Stuck~10 & 545 & 878 & 1197 & 11931  & 1028 & 984 & 1963 & 1368 & 975 & 9312  \\ \hline
        Average & \textbf{414}  & \textbf{247} & 556  & 4496  & 825  & 333 & 755  & 598 & 465  & 3569   \\ \hline
        Improvement & / &  / & 25.6$\%$  & 94.51$\%$  & 49.88$\%$  & 25.88$\%$  & 45.23$\%$  & 58.79$\%$  & 12.35$\%$  & 93.09$\%$   \\ \hline
    \end{tabular}}
\end{minipage}
\end{table}
\end{center}


\subsection{Evaluation on Coverage Rate} 

To further compare the proposed CDSL with the structural algorithm, we construct a two-stage ATPG framework on transition faults. (i) In the first stage, we set a relatively small backtrack limit and close the conflict-driven modules. We aim at handling the easy-to-detect faults with a relatively small aborted limit (The aborted limit is set $20$). (ii) In the second stage, we set a relatively large aborted limit and the proposed CDSL algorithm targets the aborted faults (The aborted limit is set at $100$). There are two baselines in the following experiments: (1) The first baseline is the one-stage conventional D-algorithm. (2) The second is also a two-stage algorithm, but the conflict-driven modules are closed in both two stages. The results are shown in Table~\ref{table:conflict-two-stage}.

It is found that the one-stage conventional D-algorithm results in $8702$ aborted faults on average, and the fault coverage rate is $92.95\%$. However, when the same D-algorithm is armed with a two-stage setting, the aborted fault can be decreased to $5975$ and the fault coverage rate can reach $95.21\%$. Further, when the proposed CDSL is implemented with a two-stage setting, aborted faults can be decreased to $4667$, and the fault coverage rate can be increased to $96.14\%$. In other words, compared with the D-algorithm, the aborted faults can be decreased via $46.37\%$ and the fault coverage rate can be increased via $3.19\%$, while compared with the two-stage algorithm which is without conflict-driven modules, the aborted faults can be decreased via $21.89\%$ and the fault coverage rate is increased via $0.93\%$. 

\begin{center}
\begin{table}[htbp]
\centering
\vspace{-0.2cm}
\begin{minipage}{7.2cm}
\centering
\caption{Evaluation in a Two-Stage Framework }
\label{table:conflict-two-stage}
\def\arraystretch{1.5}\tabcolsep 2pt
\resizebox{\textwidth}{32mm}{
   \begin{tabular}{|l|l|l|l|l|l|l|}
    \hline
\multirow{2}{*}{Circuit}&  \multicolumn{2}{|c|}{ One-Stage} & \multicolumn{2}{|c|}{ without Conflict }& \multicolumn{2}{|c|}{ Prop Model}\\
\cline{2-7}  
& UO & coverage & UO &  coverage & UO &  coverage \\ \hline
Tran~1 & 505 &  95.57\%  & 402  & 96.785\%  & 353  &  97.149\%  \\ \hline
Tran~2 & 32319 &  98.71\%  & 22710   & 99.109\%  & 17154  &  99.325\%  \\ \hline
Tran~3 & 105 &  97.86\%  & 119  &  98.867\%  & 98  &  99.029\%  \\ \hline
Tran~4 & 604 & 97.59\%  & 320  & 98.611\%  & 214  &  98.928\%  \\ \hline
Tran~5 & 5414 &  91.71\%  & 3769  &  94.678\%  & 2943  &  95.795\%  \\ \hline
Tran~6 & 13211 &  90.55\%  & 9110  &  93.548\%  & 7339  &  94.777\%  \\ \hline
Tran~7 & 14037 & 90.15\%  & 9462  &  93.383\%  & 7615  &  94.634\%  \\ \hline
Tran~8 & 13436 &  90.50\%  & 9152  &  93.603\%  & 7364  &  94.819\%  \\ \hline
Tran~9 & 1641 &  88.34\%  & 671  &  91.342\%  & 526  &  93.011\%  \\ \hline
Tran~10 & 5757 &  88.53\%  & 4043  &  92.25\%  & 3067  &  93.97\%  \\ \hline
Average & 8702  &  92.95\%  & 5975 &  95.21\% & \textbf{4667}   & \textbf{96.14}\%  \\ \hline
Improvement & 46.37\%  & 3.19\%  & 21.89\% & 0.93\% & / & / \\ \hline
\end{tabular} }
\end{minipage}
\end{table}
\end{center}


\subsection{Evaluation on Conflict Diagnosis}

Finally, we evaluate the conflict diagnosis in the case of low coverage analysis. As described in Section~\ref{subsec:diagnosis}, according to the accumulated learnt conflicts, we first mark the top $5$ logic gates. After tracing the circuits from the labeled logic gates, the conflict-related PI nodes are found, and the corresponding logic value is marked as $\alpha_N$ (supposing that there are $N$ related PI nodes). If there exist constraints on the found PI nodes, we would relax such constraints. Otherwise, if there are not any constraints on one of the found PI nodes, we prefer to add a constraint on this node and the logic value is the opposite of $\alpha$. Finally, we recall the ATPG engine to generate the test pattern or prove the untestability. The results are given in Table~\ref{table:conflict-diagnosis}. It is shown that after the conflict diagnosis, the aborted faults decrease $8.89\%$ on average, while the fault coverage rates increase by $0.271\%$.

\begin{center}
\begin{table}[htbp]
    \centering
\vspace{-0.2cm}
\begin{minipage}{5.8cm}
\centering
\caption{Evaluation on Conflict Diagnosis  }
\label{table:conflict-diagnosis}
\def\arraystretch{1.5}\tabcolsep 2pt
\resizebox{\textwidth}{28mm}{
   \begin{tabular}{|l|l|l|l|l|l|}
 \hline 
Circuit & UO & Coverage & Circuit & UO & Coverage \\ \hline
       Stuck~1  & 554  & 99.120\% &  Tran~1  & 306  & 97.337\%   \\ \hline
       Stuck~2  &  522  &  99.010\% &  Tran~2  & 14928  &  99.505\%   \\ \hline
       Stuck~3  &  920  &  98.606\% &  Tran~3  &  82  &  99.210\%  \\ \hline
       Stuck~4  &  8  &  99.803\%  & Tran~4  &  126  &  98.600\% \\ \hline
      Stuck~5  &  852  &  97.679\% & Tran~5  & 2812  &  96.004\%  \\ \hline
       Stuck~6  &  35  &  99.786\% & Tran~6  & 7002  &  95.232\%  \\ \hline
       Stuck~7  &  392  &  98.938\% & Tran~7  & 7213  &  94.887\% \\ \hline
       Stuck~8  &  2356  & 96.022\% &  Tran~8  &  6579  & 94.872\%  \\ \hline
       Stuck~9  &  5910  &  95.931\% & Tran~9  &  442  &  93.859\% \\ \hline
       Stuck~10  &  3827  &  99.873\% & Tran~10 & 2913 &  93.953\%  \\ \hline 
    \end{tabular}}
    \end{minipage}
\end{table}
\end{center}

\section{Conclusions}\label{sec:conclusion}

Aiming at addressing the efficiency problem brought by the SAT-based framework but exploiting efficient heuristics of modern SAT solver, we have proposed conflict-driven structural learning (CDSL) ATPG algorithm in this paper, which allows the structural ATPG to benefit from the SAT's techniques such as conflict management and conflict-driven branching. The proposed CDSL algorithm is composed of three parts: (1) Learnt conflict constraints before each backtrack has been constructed, aiming to learn from the mistakes and utilize the optimization process data to prune search space. (2) Conflict-driven implication and justification have been applied for decisions and implications, aiming to further increase the solving efficiency and decision effectiveness. (3) Conflict diagnosis based on the analysis of the learnt conflicts has been attempted to improve test and fault coverage rate by relaxing some of the external ATPG constraints. 
Extensive experimental results on industrial circuits have demonstrated the advantage of the proposed CDSL ATPG algorithm in three aspects: (i) Comparing with the conventional SAT-based ATPG and structural D-algorithm, the proposed CDSL algorithm has decreased the aborted faults by $25.6\%$ and $49.88\%$ on average, while the run time is decreased by $94.51\%$ and $25.88\%$, respectively. (ii) With a two-stage setting, compared with the D-algorithm, the aborted faults can be decreased via $46.37\%$ and the fault coverage rate can be increased via $3.19\%$, while compared with the two-stage algorithm which is without conflict-driven modules, the aborted faults can be decreased via $21.89\%$ and fault coverage rate is increased via $0.93\%$. (iii) Conflict diagnosis has been shown to decrease the aborted faults via $8.89\%$ on average while increasing the fault coverage rate $0.271\%$. 
Future work includes the development of more SAT heuristics on structural ATPG heuristics.

\bibliographystyle{IEEEtran}
\bibliography{refbib}

\end{document}